%% file: MICCAI.tex
\documentclass[runningheads]{llncs}
\usepackage[T1]{fontenc}
\usepackage{booktabs}
\usepackage{bbding}
\usepackage{amsmath,amssymb,amsfonts}
\usepackage{relsize}
\usepackage{makecell}
\usepackage{hyperref}
\usepackage{graphicx,verbatim}
\usepackage{color}
\usepackage{multirow}

\begin{document}
\title{Memory-Augmented Incomplete Multimodal Survival Prediction via Cross-Slide and Gene-Attentive Hypergraph Learning}
%

\author{Mingcheng Qu\inst{1} \and Guang Yang\inst{1} \and Donglin Di\inst{2} \and Yue Gao\inst{2} \and Tonghua Su\inst{1} \and Yang Song\inst{3} \and Lei Fan\inst{3}\Envelope
}  
\authorrunning{Mingcheng Qu et al.}
\titlerunning{M²Surv}
\institute{$^1$Faculty of Computing, Harbin Institute of Technology \\
     $^2$School of Software, Tsinghua University\\
$^3$School of Computer Science and Engineering, UNSW Sydney\\
    \email{lei.fan1@unsw.edu.au}}

\maketitle              
\begingroup                      
  \renewcommand\thefootnote{\Envelope}
  \footnotetext{Corresponding author}
\endgroup

\begin{abstract}
    Multimodal pathology-genomic analysis is critical for cancer survival prediction. However, existing approaches predominantly integrate formalin-fixed paraffin-embedded (FFPE) slides with genomic data, while neglecting the availability of other preservation slides, such as  Fresh Froze (FF) slides. Moreover, as the high-resolution spatial nature of pathology data tends to dominate the cross-modality fusion process, it hinders effective multimodal fusion and leads to modality imbalance challenges between pathology and genomics. These methods also typically require complete data modalities, limiting their clinical applicability with incomplete modalities, such as missing either pathology or genomic data. 
    In this paper, we propose a multimodal survival prediction framework that leverages hypergraph learning to effectively integrate multi-WSI information and cross-modality interactions between pathology slides and genomics data while addressing modality imbalance. In addition, we introduce a memory mechanism that stores previously learned paired pathology-genomic features and dynamically compensates for incomplete modalities. Experiments on five TCGA datasets demonstrate that our model outperforms advanced methods by over 2.3\% in C-Index. Under incomplete modality scenarios, our approach surpasses pathology-only (3.3\%) and gene-only models (7.9\%). Code: \url{https://github.com/MCPathology/M2Surv}



\end{abstract}

\input{chapter/Introduction}

\input{chapter/Method}
\input{chapter/Experiment}
\input{chapter/Conclusion}

\bibliographystyle{splncs04}
\bibliography{miccai25}

\end{document}

%% file: chapter/Introduction.tex
\section{Introduction}
Multimodal survival prediction, integrating Whole Slide Images (WSIs) with genomic profiles, offers great potential for advancing precision oncology~\cite{nunes2024prognostic,zhu2025interpretable}. This integration leverages the complementary strengths: WSIs capture cellular morphology and tumor micro-environment~\cite{zhang2019pathologist,tang2025prototype}, while genomic profiles identify key driver mutations and define molecular subtypes~\cite{qu2024boundary,qu2025multimodal}. 

Generally, two prevalent methods are used for WSI preparation: Formalin-Fixed Paraffin-Embedded (FFPE) and Fresh Frozen (FF). Specifically, FFPE slides are widely used due to their high-quality morphological preservation, while FF slides offer better nucleic acids and proteins but are more prone to artifacts and structural degradation. For patients with multiple slides, previous studies~\cite{Mcat,zhang2024prototypical,CMTA,SurvPath} typically aggregated features from all patches across different slides, overlooking the \textbf{heterogeneity} in staining style within WSIs ~\cite{fan2022cancer}. 

On the other hand, these models employ mid- or late-feature fusion strategies for multimodal integration of pathology and genomics, achieving better performance compared to unimodal approaches~\cite{Mcat,MOTCat,SurvPath}. However, they face challenges related to \textbf{modality imbalance}, as WSIs contain thousands of patches while only a few hundred genes are identified for common cancers~\cite{raser2005noise}. This imbalance leads to the pathology modality dominating the fusion process, particularly when using cross-attention mechanisms~\cite{Mcat,SurvPath}. Incorporating slides of both FFPE and FF types would further exacerbate this issue.
Furthermore, in practice, technical and financial constraints often result in insufficient tissue samples and sequencing errors~\cite{Dorent2019HeteroModalVE,shen2019brain}, leading to \textbf{incomplete modalities}, such as missing genomic data or pathology WSIs. However, the effectiveness of multimodal fusion strategy (\textit{e.g.}, cross-attention) depends heavily on complete correlations between modalities, which limits their clinical applicability and poses deployment challenges in real-world settings.


In this paper, we propose M²Surv, a \textbf{M}emory-augmented Incomplete \textbf{M}ulti-modal \textbf{Sur}vival Prediction Framework, consisting of three stages: feature extraction, multi-slide hypergraph, and gene-attentive hypergraph. 
Specifically, the multi-slide hypergraph represents multiple pathology slides by treating patches as nodes, and first constructs intra-WSI hyperedges within individual slides based on spatial interactions, then progressively aggregates information across multiple slides through inter-WSI hyperedges, capturing morphological variations and histological patterns at different levels. To mitigate the pathology-genomics imbalance, the gene-attentive hypergraph establishes dense cross-modal connections by explicitly linking each gene group to all pathology features. By doing this, the importance of genomic features is reinforced, ensuring a more balanced contribution to the fusion process. Additionally, a memory bank is introduced to store paired pathology-genomic features during training, allowing the retrieval of relevant features to compensate for incomplete modalities during inference, ensuring reliable predictions even with incomplete data.

Our contributions are summarized as: A multimodal framework, M²Surv, is proposed to integrate multiple pathology slides and genomic data while addressing the pathology-genomics imbalance through hypergraph learning. A memory bank is implemented with few computational consumption yet effectively compensates for missing modality. Extensive experiments demonstrate the superiority of our model, achieving advanced performance across five TCGA datasets.

%% file: chapter/Method.tex
\section{Method}

\begin{figure}[t]
    \centering
    \includegraphics[width=1\linewidth]{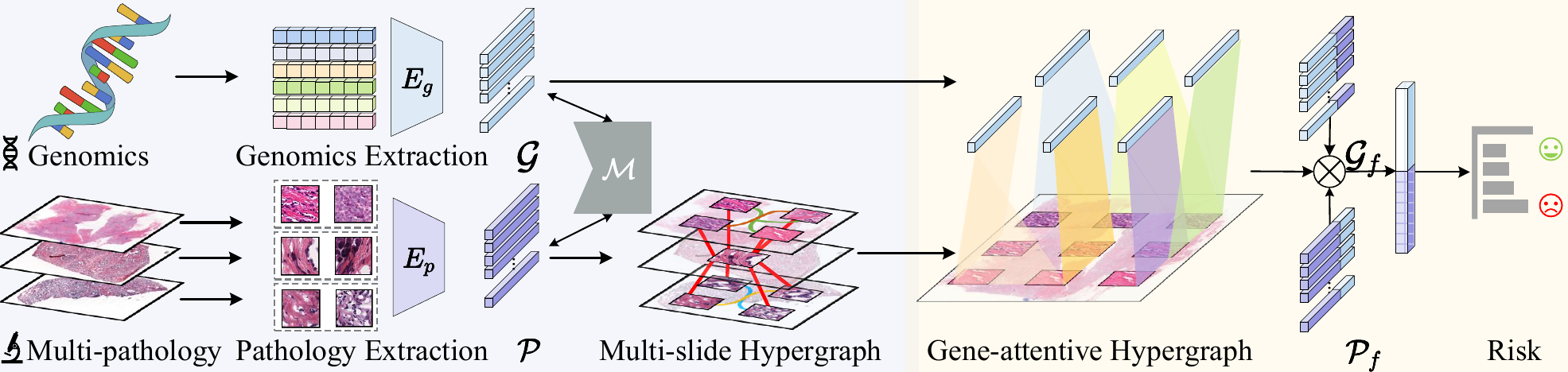}
    \caption{\textbf{Overview of M²Surv.} It includes \textit{feature extraction} to process multiple pathology slides and genomics, \textit{multi-slide hypergraph} to capture both intra-slide and inter-slide feature representations, and \textit{gene-attentive hypergraph} to establish dense connections between each gene group and all pathology patches. \textit{A memory bank} is incorporated to store paired pathology-genomic features during training and retrieve similar features during inference, mitigating missing modality challenges.}
    \label{fig:model}
\end{figure}

\subsection{Overview}
\textbf{Preliminary.} Given a cohort \( \mathbb{X} = \{\mathrm{X}_1, \dots, \mathrm{X}_n\} \) of \( n \) subjects, each subject \( \mathrm{X}_i = \{\mathrm{H}_i, y_i\} \) consists of pathology–genomics features \( \mathrm{H}_i = \{\mathcal{P}_i, \mathcal{G}_i\} \) and survival information \( y_i = \{c_i, t_i\} \). Here, \( \mathcal{P}_i = \{\mathrm{P}_{i1}, \dots, \mathrm{P}_{iK}\} \) includes \( K \) FF and FFPE slides, \( \mathcal{G}_i \) represents the genomic profiles, \( c_i \in \{0,1\} \) denotes the event status (\( c_i = 0 \) indicates event occurrence), and \( t_i \) is the overall survival time.
The goal is to estimate the hazard function $\phi_{h}(t)$, which predicts the instantaneous incidence event rate at time $t$, while training a model $\mathcal{F}$ to predict the probability of survival beyond $t$ using the survival function $\phi_{s}(t)$. The model is optimized using the negative log-likelihood losses~\cite{YAO2020101789}, defined as:
{\small
\begin{equation}
    \mathcal{L}_{\mathcal{S}} =
\textstyle\sum_{i=1}^{n}(1-c_i)\log{\phi_h(t_i|\mathrm{H_i})}   
+ c_i\log{\phi_s(t_i|\mathrm{H_i})} +(1-c_i)\log{\phi_{s}(t_i-1|\mathrm{H_i})}.
\end{equation}
}
\textbf{Our Framework.}
M²Surv comprises feature extraction, multi-slide hypergraph, and gene-attentive hypergraph (see Fig.~\ref{fig:model}). 
Paired pathology \( \mathcal{P} \) and genomics \( \mathcal{G} \) features are extracted using respective encoders. The multi-slide hypergraph first builds intra-WSI graphs for each slide based on spatial interactions, followed by an inter-WSI hypergraph to capture structural relationships across multiple slides. The gene-attentive hypergraph constructs dense connections using gene features to guide and refine \( \mathcal{P}_h \), generating integrated features \( \mathcal{P}_f \) and \( \mathcal{G}_f \) for final risk prediction.
To handle incomplete modalities, a memory bank \( \mathcal{M} \) is introduced to store paired pathology-genomic features \( \mathrm{M}_p \) and \( \mathrm{M}_g \) during training, allowing the model to retrieve and approximate the similar features from \( \mathcal{M} \) to compensate for missing information during inference.

\textbf{Feature Extraction.}
Following previous studies~\cite{CLAM,SurvPath,zhang2024prototypical}, each WSI \(\mathrm{P}_k\) is partitioned and randomly selected into $N_k=4096$ patches of $256 \times 256$ pixel, at 20\(\times\) magnification. A pretrained encoder (ResNet50) extracts \(d\)-dimensional features from these patches, representing the WSI as \(\mathrm{P}_k \in \mathbb{R}^{N_k \times d} = \{p_1, \ldots, p_{N_k}\}\), where each patch \(p_k\) has spatial coordinates \(\zeta_{p_k} = (x_k, y_k)\). The multi-pathology feature set \(\mathcal{P} = \{\mathrm{P}_1, \ldots, \mathrm{P}_K\}\) includes multiple slide features from the same patient.
For genomic data like RNA-seq, CNV, and SNV, we adopt the feature selection method~\cite{Mcat} by grouping them into $W=6$ functional groups: Tumor Suppression, Oncogenesis, Kinases, Cellular Differentiation, Transcription, and Cytokines. Each category is encoded using a genomic encoder (\textit{i.e.}, a multilayer perceptron, MLP) to produce genomic features \(\mathcal{G} \in \mathbb{R}^{W \times d} = \{g_1, \ldots, g_W\}\).

\subsection{Multi-slide Hypergraph}
Considering the distinct staining styles and biopsy tissue variations in multiple slides, we adopt a two-stage strategy: intra-slide to handle style differences within individual slides, followed by inter-slide integration to ensure the aggregation of multi-slide information. Specifically, we leverage widely used hypergraphs~\cite{di2022big,di2022generating,jing2025multi,yi2022approximate} to construct \textbf{intra-slide topological hyperedges} for each WSI while establishing \textbf{inter-slide structural hyperedges} across different slides.

For a WSI \( \mathrm{P}_k \), each patch \( p \) is treated as a vertex, and intra-slide hyperedges are formed by grouping each patch with its neighboring patches based on Euclidean distance. Given a patch \( p_k \), its neighbors are determined as:  
\begin{equation}
\mathcal{N}_T (p_k) = \{p_j \ | \ \zeta_{p_j} - \zeta_{p_k}\|_2 \leq \delta\},
\end{equation}
where \( \zeta_{p_j} \) and \( \zeta_{p_k} \) denote the coordinates of patches \( p_j \) and \( p_k \) respectively, and \( \delta \) is a distance threshold. This yields topological-based hyperedges: \(\mathcal{E}_T^{(k)} = \{\{p_k, p_{j1}, p_{j2}, \dots\} \ | \ \forall p_j \in N_T (p_k)\}\).  The combined intra-slide hyperedges across \( K \) WSIs are then represented as \( \mathcal{E}_T = \{\mathcal{E}_T^{(1)}, \dots, \mathcal{E}_T^{(K)}\} \).
By capturing the topological relationships between patches, this approach effectively encodes spatial structures within each WSI, thereby preserving the inherent tissue morphology.

For multi-pathology $\mathcal{P}$, all patches are treated as nodes, and inter-slide structural hyperedges using feature similarity. The neighbors of $p_k$ are identified as:
\begin{equation}
    \mathcal{N}_F(p_k) = \{p_j \mid \text{sim}(p_k, p_j) \geq \alpha \},
\end{equation}
where $\text{sim}(\cdot, \cdot)$ represents the cosine similarity function, and $\alpha$ is a similarity threshold. This builds structural hyperedges: $\mathcal{E}_F = \{\{p_k, p_{j_1}, p_{j_2}, \dots \} \mid \forall~ p_j \in \mathcal{N}_F(p_k) \}.$
By capturing the structural relationships between patches across multiple WSIs, it identifies shared morphological patterns and preserves consistent biological information across WSIs. Notably, \(\delta\) and \(\alpha\) are determined by the hyperedge construction threshold \(\lambda\), which selects \(\lambda-1\) neighbors for each patch. 
 
The final hyperedge set $ \mathcal{E}_m $ is formed by merging these intra-slide and inter-slide hyperedges, where $\mathcal{E}_m = \mathcal{E}_T \cup \mathcal{E}_F$ represents their union. The constructed multi-slide hypergraph is then processed through hypergraph convolutions~\cite{feng2019hypergraph} to extract high-order feature representations \(\mathcal{P}_h = \mathcal{P}^{(L)}\) after \(L\) layers.



\subsection{Gene-attentive Hypergraph}
Considering the dimensional imbalance between genomic and pathology features, we aim to construct a dense gene-attentive hypergraph by establishing connections between each gene group and the multi-slide hypergraph. It is motivated by the ability of hypergraphs to link features from different modalities into interconnected clusters, effectively capturing complex cross-modal relationships within a unified structure.
Specifically, all nodes in the multi-slide hypergraph and the six gene groups are treated as new nodes, with hyperedges $\mathcal{E}_g$ centered around each gene group. The neighbors are identified as:
\begin{equation}
    \mathcal{N}_G(g_w)=\{p_j\mid\frac{\exp(\text{att}(g_w, p_j))}{\sum_{k=1}^{N} \exp(\text{att}(g_w, p_k))}\geq\beta\},
\end{equation}
where $att$ represents the cross-attention score between gene group $g_w$ and patch $p_j$, and $\beta$ is the threshold (empirically set to select the top 5\% of patch-gene hyperedges balancing connections and computation). This results gene-attentive hyperedges: $\mathcal{E}_G = \{\{g_w, p_{j_1},$ $ p_{j_2}, \dots \} \mid \forall~ p_j \in \mathcal{N}_G(g_w) \}$. Then, the hypergraph convolution is employed to update the representations of all nodes and produce the refined pathology and genomic features \( \mathcal{P}_f \) and \( \mathcal{G}_f \).  

The gene-attentive hypergraph leverages cross-attention mechanisms to define hyperedges, emphasizing cross-modal relationships and interactions. By establishing dense connections centered on gene groups, our approach enables each gene group to effectively interact with multiple pathology patches, mitigating the imbalance issue observed in previous cross-attention strategies~\cite{Mcat,SurvPath}.


\subsection{Memory Bank}
To address incomplete modalities in clinical scenarios, we introduce a memory bank $\mathcal{M}$~\cite{alonso2021semi} to store paired pathology-genomic features $ \{\left<\mathcal{P},\mathcal{G}\right>\}^{n}_{i=1}$ during training, where $n$ is the number of training samples.
We employ a momentum update strategy~\cite{he2020momentum} to dynamically update stored features. At the training $r$- epoch, $\mathcal{M}$ is updated: $\mathcal{M}^{(r)} \leftarrow  \theta \cdot \left<\mathcal{P}^{(r)},\mathcal{G}^{(r)}\right> + (1-\theta) \cdot \mathcal{M}^{(r-1)}$, where $\theta\in [0,1]$ is the momentum coefficient. 

During inference, if a modality (\textit{i.e.}, all pathology slides or genomic data) is missing, the available modality \(\mathrm{M_a}\) is used to retrieve the most relevant features by computing its cosine similarity with all entries in \(\mathcal{M}\). The top-$\mu$ most similar entries \(\{\mathrm{M}^{(j)}_{m}\}^{\mu}_{j=1}\) are selected and then aggregated, expressed as:
\begin{equation}
    \mathrm{\hat{M}}_m = \sum_{j=1}^\mu \frac{\exp(\text{sim}(\mathrm{M}_a,\mathrm{M}_a^{(j)}))}{\sum^{\mu}_{l=1}\exp(\text{sim}(\mathrm{M}_a,\mathrm{M}^{(l)}_a))}\mathrm{M}^{(j)}_{m},
\end{equation}
where \(\mathrm{\hat{M}}_m\) is the approximated features used to represent the missing features.

Our memory bank leverages momentum update retrieval for collective analysis of historical training data. Essentially, it integrates multimodal information from previously learned multiple samples, effectively addressing incomplete modalities while maintaining efficiency.

%% file: chapter/Experiment.tex
\section{Experiments}

\begin{table*}[t]
\caption{\textbf{Comparison of our model with advanced methods on five datasets.} C-Indexes (Mean $\pm$ STD) are reported based on 5-fold cross-validation.}
\label{table:comparison-merged}
\centering
\resizebox{\textwidth}{!}{
\begin{tabular}{ccl|ccccc|c}
    \toprule
 \multicolumn{3}{c|}{\textbf{Model}} &\textbf{BLCA} & \textbf{BRCA} & \textbf{CO-READ} & \textbf{HNSC} & \textbf{STAD} & \textbf{Mean} \\
    \toprule
    
    \multicolumn{2}{c}{\multirow{5}{*}{\rotatebox[origin=c]{90}{\textbf{Pathology}}}}& ABMIL~\cite{ilse2018attention}  & 0.624 \textsmaller{$\pm$ 0.059} & 0.672 \textsmaller{$\pm$ 0.051} & 0.730 \textsmaller{$\pm$ 0.151} & 0.624 \textsmaller{$\pm$ 0.042} & 0.636 \textsmaller{$\pm$ 0.043} & 0.657 \\ 
    && AMISL~\cite{YAO2020101789} & 0.627 \textsmaller{$\pm$ 0.032} & 0.681 \textsmaller{$\pm$ 0.036} & 0.710 \textsmaller{$\pm$ 0.091} & 0.607 \textsmaller{$\pm$ 0.048} & 0.553 \textsmaller{$\pm$ 0.012} & 0.636 \\ 
    &&TranMIL~\cite{shao2021transmil}   & 0.617 \textsmaller{$\pm$ 0.045} & 0.663 \textsmaller{$\pm$ 0.053} & 0.747 \textsmaller{$\pm$ 0.151} & 0.619 \textsmaller{$\pm$ 0.062} & 0.660 \textsmaller{$\pm$ 0.072} & 0.661 \\ 
    &&CLAM-MB~\cite{CLAM}   & 0.623 \textsmaller{$\pm$ 0.045} & 0.696 \textsmaller{$\pm$ 0.098} & 0.721 \textsmaller{$\pm$ 0.159} & \textbf{0.648 \textsmaller{$\pm$ 0.050}} & 0.620 \textsmaller{$\pm$ 0.034} & 0.662 \\ 
    &&M²Surv (\textbf{Ours})& \textbf{0.646 \textsmaller{$\pm$ 0.022}} & \textbf{0.735 \textsmaller{$\pm$ 0.056}} & \textbf{0.749 \textsmaller{$\pm$ 0.045}} & 0.612 \textsmaller{$\pm$ 0.004} & \textbf{0.677 \textsmaller{$\pm$ 0.048}} & \textbf{0.684} \\
    \cline{1-9}
    \multicolumn{2}{c}{\multirow{4}{*}{\rotatebox[origin=c]{90}{\textbf{Genomic}}}}&MLP~\cite{haykin1998neural}  &   0.530 \textsmaller{$\pm$ 0.070} & 0.622 \textsmaller{$\pm$ 0.079} & 0.712 \textsmaller{$\pm$ 0.114} & 0.520 \textsmaller{$\pm$ 0.064} & 0.497 \textsmaller{$\pm$ 0.031} & 0.576 \\ 
    &&SNN~\cite{klambauer2017self}  &   0.521 \textsmaller{$\pm$ 0.070} & 0.621 \textsmaller{$\pm$ 0.073} & 0.711 \textsmaller{$\pm$ 0.162} & 0.514 \textsmaller{$\pm$ 0.076} & 0.485 \textsmaller{$\pm$ 0.047} & 0.570 \\ 
    &&SNNTrans~\cite{klambauer2017self}  &   0.583 \textsmaller{$\pm$ 0.060} & 0.679 \textsmaller{$\pm$ 0.053} & \textbf{0.739 \textsmaller{$\pm$ 0.124}} & 0.570 \textsmaller{$\pm$ 0.035} & 0.547 \textsmaller{$\pm$ 0.041} & 0.622 \\ 
     &&M²Surv (\textbf{Ours}) & \textbf{0.593 \textsmaller{$\pm$ 0.065}} & \textbf{0.696 \textsmaller{$\pm$ 0.043}} & 0.701 \textsmaller{$\pm$ 0.087} & \textbf{0.659 \textsmaller{$\pm$ 0.058}} & \textbf{0.704 \textsmaller{$\pm$ 0.082}} & \textbf{0.671} \\
     \cline{1-9}
    \multicolumn{2}{c}{\multirow{8}{*}{\rotatebox[origin=c]{90}{\textbf{Multimodal}}}}&SNN+CLAM & 0.625 \textsmaller{$\pm$ 0.060} & 0.699 \textsmaller{$\pm$ 0.064} & 0.716 \textsmaller{$\pm$ 0.016} & 0.638 \textsmaller{$\pm$ 0.066} & 0.629 \textsmaller{$\pm$ 0.065} & 0.661 \\ 
    &&Porpoise~\cite{chen2022pan}  & 0.617 \textsmaller{$\pm$ 0.056} & 0.668 \textsmaller{$\pm$ 0.070} & 0.738 \textsmaller{$\pm$ 0.151} & 0.614 \textsmaller{$\pm$ 0.058} & 0.660 \textsmaller{$\pm$ 0.106} & 0.659 \\ 
    &&MCAT~\cite{Mcat} & 0.640 \textsmaller{$\pm$ 0.076} & 0.685 \textsmaller{$\pm$ 0.109} & 0.724 \textsmaller{$\pm$ 0.137} & 0.564 \textsmaller{$\pm$ 0.084} & 0.625 \textsmaller{$\pm$ 0.118} & 0.648 \\ 
    &&MOTCat~\cite{MOTCat} & 0.659 \textsmaller{$\pm$ 0.069} & 0.727 \textsmaller{$\pm$ 0.027} & 0.742 \textsmaller{$\pm$ 0.124} & 0.656 \textsmaller{$\pm$ 0.041} & 0.621 \textsmaller{$\pm$ 0.065} & 0.681 \\ 
    &&CMTA~\cite{CMTA} &  0.670 \textsmaller{$\pm$ 0.030} & 0.691 \textsmaller{$\pm$ 0.037} & 0.704 \textsmaller{$\pm$ 0.117} & 0.562 \textsmaller{$\pm$ 0.086} & 0.592 \textsmaller{$\pm$ 0.014} & 0.644 \\ 
    &&SurvPath~\cite{SurvPath}  & 0.635 \textsmaller{$\pm$ 0.026} & 0.679 \textsmaller{$\pm$ 0.077} & 0.731 \textsmaller{$\pm$ 0.124} & 0.617 \textsmaller{$\pm$ 0.058} & 0.620 \textsmaller{$\pm$ 0.044} & 0.656 \\ 
    &&PIBD~\cite{zhang2024prototypical} & 0.651 \textsmaller{$\pm$ 0.092} & 0.712 \textsmaller{$\pm$ 0.048} &\textbf{0.786 \textsmaller{$\pm$ 0.134}} & 0.607 \textsmaller{$\pm$ 0.059} & 0.668 \textsmaller{$\pm$ 0.055} & 0.685 \\ 
    
    &&M²Surv (\textbf{Ours}) & \textbf{0.671 \textsmaller{$\pm$ 0.039}} & \textbf{0.744 \textsmaller{$\pm$ 0.091}} & 0.757 \textsmaller{$\pm$ 0.080} & \textbf{0.661 \textsmaller{$\pm$ 0.012}} & \textbf{0.673 \textsmaller{$\pm$ 0.028}} & \textbf{0.701} \\ 
    \bottomrule
\end{tabular}
}
\end{table*}

\textbf{Datasets and Experimental Settings.} Following previous studies~\cite{SurvPath,zhang2024prototypical}, we evaluated our models on five The Cancer Genome Atlas (TCGA) datasets: Bladder Urothelial Carcinoma (BLCA) (n=384), Breast Invasive Carcinoma (BRCA) (n=968), Colon and Rectum Adenocarcinoma (CO-READ) (n=298), Head and Neck Squamous Cell Carcinoma (HNSC) (n=392), and Stomach Adenocarcinoma (STAD) (n=317). We followed the previous dataset settings~\cite{Mcat,SurvPath,zhang2024prototypical}, and employed the Adam optimizer with a learning rate of $1 \times 10^{-4}$ and a weight decay of $1 \times 10^{-5}$, training 30 epochs. Concordance Index (C-Index) was used as the metric. For each dataset, We performed 5-fold cross-validation with a 4:1 train-val split, reporting results as the mean $\pm$ standard deviation (STD). 



\subsection{Comparisons with Advanced Methods}
We compared our models in three settings: pathology-only (ABMIL, AMISL, TransMIL, and CLAM), genomic-only (MLP, SNN, and SNNTrans), and multimodal (Porpoise, MCAT, MOTCat, CMTA, SurvPath, and PIBD). Results for unimodal methods, SNN+CLAM, and Porpoise were quoted from previous study~\cite{zhang2024prototypical}, while others were reproduced using their released code (see Table~\ref{table:comparison-merged}).

Our model achieved a mean C-index of 0.701, outperforming all unimodal and multimodal methods on nearly all datasets.
In modality-missing scenarios, it exhibited superior performance with a mean C-Index of 0.684 (pathology-only),
surpassing TransMIL (0.661) and CLAM-MB (0.662). 
Similarly, it maintained a score of 0.671, outperforming gene-only models such as SNNTrans (0.622). These results demonstrate the effectiveness of our framework in integrating pathology and genomic data while handling incomplete modalities.

+
\begin{table}[t]
\centering
\caption{\textbf{Ablation study on multi-slide hypergraph and gene-attentive hypergraph}. We evaluated pathology aggregator ($Agg$), slide type (HGNN$^-$ for FFPE only), hyperedge types, and construction threshold $\lambda$ for multi-slide hypergraph, and multimodal feature fusion ($Fuse$) and cross-modal edge construction methods (HGNN$^*$ for random edge instead of attention score) for gene-attentive hypergraph. }
\label{tab:ablation}
\resizebox{\textwidth}{!}{
\begin{tabular}{ccccc|ccccc|c}
\toprule
&$Agg$  & Slides & $\lambda$ &$Fuse$&BLCA&BRCA&CO-READ&HNSC&STAD&Mean \\

\cmidrule(lr){1-11}

\multirow{10}{*}{\rotatebox[origin=c]{90}{\textbf{Multi-slide hypergraph}}}&MLP  & $Multi$ & 9 &HGNN &0.644 \textsmaller{$\pm$ 0.064} & 0.708 \textsmaller{$\pm$ 0.039}& 0.750 \textsmaller{$\pm$ 0.046}& 0.620 \textsmaller{$\pm$ 0.043} & 0.638 \textsmaller{$\pm$ 0.056} &  0.672\\
&ABMIL  & $Multi$ & 9 &HGNN &  0.645 \textsmaller{$\pm$ 0.034}& 0.715 \textsmaller{$\pm$ 0.101}& 0.699 \textsmaller{$\pm$ 0.049} & 0.632 \textsmaller{$\pm$ 0.037} & 0.661 \textsmaller{$\pm$ 0.031} & 0.670\\
&TransMIL  & $Multi$ & 9 &HGNN &0.634 \textsmaller{$\pm$ 0.045} & 0.718 \textsmaller{$\pm$ 0.083}& 0.743 \textsmaller{$\pm$ 0.109}& 0.623 \textsmaller{$\pm$ 0.038} & 0.661 \textsmaller{$\pm$ 0.056} &  0.676\\
&GAT  & $Multi$ &9&HGNN &0.648 \textsmaller{$\pm$ 0.039} & 0.733 \textsmaller{$\pm$ 0.116}& 0.730 \textsmaller{$\pm$ 0.136}& 0.637 \textsmaller{$\pm$ 0.040} & 0.656 \textsmaller{$\pm$ 0.094} &  0.681 \\
&GCN  & $Multi$ &9 &HGNN & 0.644 \textsmaller{$\pm$ 0.038} &  0.701 \textsmaller{$\pm$ 0.053}& 0.755 \textsmaller{$\pm$ 0.086}& 0.621 \textsmaller{$\pm$ 0.020} & 0.667 \textsmaller{$\pm$ 0.044} &  0.678 \\
\cmidrule(lr){2-11}
&HGNN$^-$  & intra. &9 &HGNN  &0.668 \textsmaller{$\pm$ 0.045}& 0.733 \textsmaller{$\pm$ 0.062}& \textbf{0.767 \textsmaller{$\pm$ 0.095}}& 0.639 \textsmaller{$\pm$ 0.056} & 0.671 \textsmaller{$\pm$ 0.067} &  0.696 \\

&HGNN  &intra.  & 9 &HGNN &0.645 \textsmaller{$\pm$ 0.041}& 0.684 \textsmaller{$\pm$ 0.065}& 0.706 \textsmaller{$\pm$ 0.122}& 0.631 \textsmaller{$\pm$ 0.021} & 0.661 \textsmaller{$\pm$ 0.066} &  0.665\\
&HGNN &inter.  & 9&HGNN& 0.632 \textsmaller{$\pm$ 0.024}& 0.680 \textsmaller{$\pm$ 0.103}& 0.742 \textsmaller{$\pm$ 0.062}& 0.627 \textsmaller{$\pm$ 0.039} & 0.639 \textsmaller{$\pm$ 0.079} &  0.664 \\
&HGNN  & $Multi$ & 5&HGNN & 0.649 \textsmaller{$\pm$ 0.032}& 0.695 \textsmaller{$\pm$ 0.048}& 0.689 \textsmaller{$\pm$ 0.067} & 0.641 \textsmaller{$\pm$ 0.082} & \textbf{0.706 \textsmaller{$\pm$ 0.088}} &  0.676\\
&HGNN & $Multi$ & 25&HGNN& 0.640 \textsmaller{$\pm$ 0.028}& 0.734 \textsmaller{$\pm$ 0.061}& 0.709 \textsmaller{$\pm$ 0.080} & 0.620 \textsmaller{$\pm$ 0.040} & 0.660 \textsmaller{$\pm$ 0.067} &  0.673\\

\cmidrule(lr){1-11}

\multirow{5}{*}{\rotatebox[origin=c]{90}{\textbf{Gene-attn}}}&HGNN  & $Multi$ &9&Concat &0.628 \textsmaller{$\pm$ 0.049} & 0.720 \textsmaller{$\pm$ 0.035}& 0.708 \textsmaller{$\pm$ 0.026}& 0.590 \textsmaller{$\pm$ 0.065} & 0.673 \textsmaller{$\pm$ 0.048} &  0.664\\
&HGNN & $Multi$ &9&Co-Attn & 0.669 \textsmaller{$\pm$ 0.030}& 0.677 \textsmaller{$\pm$ 0.035}& 0.713 \textsmaller{$\pm$ 0.087} & 0.632 \textsmaller{$\pm$ 0.033} & 0.655 \textsmaller{$\pm$ 0.048} & 0.669\\
&HGNN  & $Multi$ &9&GAT  &  0.623 \textsmaller{$\pm$ 0.039} & 0.731 \textsmaller{$\pm$ 0.089} & 0.717 \textsmaller{$\pm$ 0.126} & 0.642 \textsmaller{$\pm$ 0.070} & 0.651 \textsmaller{$\pm$ 0.059} & 0.673 \\
&HGNN  &  $Multi$ &9&GCN  & 0.646 \textsmaller{$\pm$ 0.037} & 0.732 \textsmaller{$\pm$ 0.095} & 0.692 \textsmaller{$\pm$ 0.097} & 0.607 \textsmaller{$\pm$ 0.059} & 0.703 \textsmaller{$\pm$ 0.091} & 0.676 \\
&HGNN  & $Multi$ &9&HGNN$^*$  & 0.601 \textsmaller{$\pm$ 0.011} & 0.702 \textsmaller{$\pm$ 0.066} & 0.686 \textsmaller{$\pm$ 0.104} & 0.629 \textsmaller{$\pm$ 0.063} & 0.654 \textsmaller{$\pm$ 0.030} & 0.654 \\

\cmidrule(lr){1-11}
&HGNN & $Multi$ &9 &HGNN & \textbf{0.671 \textsmaller{$\pm$ 0.039}} & \textbf{0.744 \textsmaller{$\pm$ 0.091}} & 0.757 \textsmaller{$\pm$ 0.080} & \textbf{0.661 \textsmaller{$\pm$ 0.012}} & 0.673 \textsmaller{$\pm$ 0.028} & \textbf{0.701} \\
\bottomrule
\end{tabular}
}
\end{table}

\subsection{Ablation Studies}
\textbf{Multi-slide hypergraph.} We evaluated various pathology aggregators ($Agg$), slide types, hyperedge types, and hyperedge construction thresholds $\lambda$ (see Table~\ref{tab:ablation}). 
When using $Agg$ such as MLP (0.672), ABMIL \cite{ilse2018attention} (0.670), and TransMIL \cite{shao2021transmil} (0.676), incorporating GAT \cite{GAT}, GCN \cite{GCN} and HGNN improved the average C-Index to 0.681, 0.698 and 0.701 respectively. highlighting the effectiveness of graph-based models in enhancing feature aggregation. Using only FFPE (HGNN$^-$) yielded a score of 0.696, which increased to 0.701 with multi-slide integration, demonstrating the benefit of incorporating multiple slides.


When exploring different slide types, using only intra-slide topological hyperedges or inter-slide structural hyperedges achieved a mean C-Indexes of 0.665 and 0.664 respectively, indicating that both connections contribute to model performance. 
Moreover, varying HGNN construction thresholds (5, 9, and 25) produced scores of 0.676, 0.701, and 0.673, respectively. This suggests that an optimal threshold balances information retention, preventing homogenization at ($\lambda$ = 25) and avoiding oversimplification at ($\lambda$ = 5).

\begin{figure}[t]
    \centering
    \includegraphics[width=1.0\linewidth]{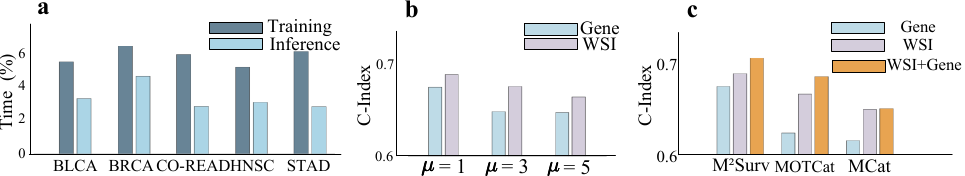}
    \caption{
    \textbf{a)}, Extra time consumption for memory bank during training and inference. \textbf{b)}, Performance with varying retrieval top-$\mu$. \textbf{c)}, Performance with incomplete modality using the memory bank across models (M²Surv, MCat,and MOTCat). 
    }
    \label{fig:res}
\end{figure}

\textbf{Gene-attentive hypergraph.}
We evaluated multimodal fusion and hyperedge construction methods (see Table \ref{tab:ablation}). Compared to direct concatenation (0.664) and cross-attention (0.669), gene-attentive GAT and GCN improved the C-Index to 0.673 and 0.676, respectively. 
HGNN with attention-based edge construction achieved a score of 0.701 better than random edge construction (0.654), confirming the effectiveness of hypergraph integration with attention mechanisms in capturing cross-modal interactions and mitigating modality imbalance.

\textbf{Memory bank.}
We evaluated the time consumption, top-$\mu$, and generalizability of our memory bank across five datasets (see Fig.~\ref{fig:res}). 
Specifically, using the memory bank across various datasets resulted in an approximately 4.5\% increase in overall time consumption and a 2.5\% in inference time.
When selecting the top \(\mu\) relevant features, the best average performance was achieved at \(\mu = 1\), indicating that utilizing only the most relevant stored feature is effective in mitigating missing modalities.
Incorporating the memory bank into MCat~\cite{Mcat} and MOTCat~\cite{MOTCat} enabled these models to perform effectively even with incomplete modalities while maintaining performance comparable to full settings, showing the adaptability of the memory bank in enhancing multimodal methods.


\begin{figure}[t]
    \centering
    \includegraphics[width=1\linewidth]{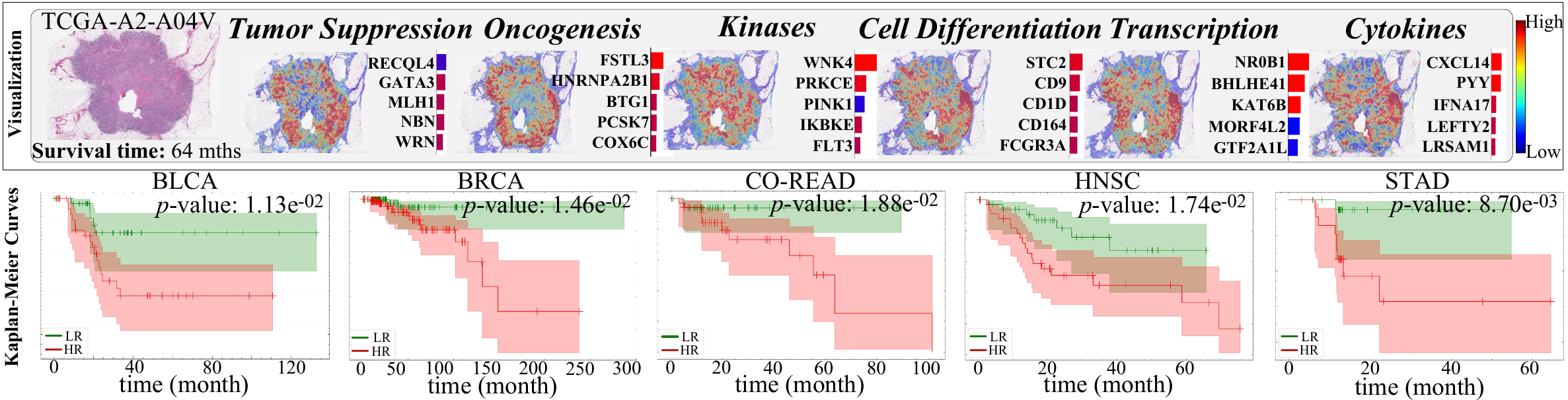}
    \caption{
    \textbf{Visualization for BRCA cases:} The heatmap generated from cross-attention scores, and the top five most influential genes are highlighted by gradient integral.
    \textbf{Kaplan-Meier curves (Bottom)} shows significant survival stratification (p<0.05 across all datasets) between high/low-risk groups (median split).
    }
    \label{fig:visualization}
\end{figure}

\subsection{Visualization}
The attention score visualizations highlight specific regions targeted by different gene groups, quantitatively assessing the gene-attentive hypergraph in capturing cross-modal interactions (see Fig.~\ref{fig:visualization}). For example, it reveals individual gene influences, with NR0B1 exhibiting a high positive gradient, indicating an enhancing role in certain pathological conditions.  
To validate the discriminative capacity of our model, we performed Kaplan-Meier analysis and log-rank test~\cite{mantel1966evaluation} by stratifying patients into high- and low-risk groups based on the predicted median risk scores. The \textit{p}-values below 0.05 confirmed the effectiveness of the model across all datasets.

%% file: chapter/Conclusion.tex
\section{Conclusion}
We proposed M²Surv, a multimodal survival prediction framework that leverages hypergraphs to model multi-slide and genomic data. We introduced a memory bank to handle missing clinical modalities. Extensive experiments demonstrated the superior performance of our framework.

\textbf{Limitations.} While incorporating more slides enhances consistency modeling, artifacts in FF slides may introduce noise into pathology features. Hypergraphs effectively capture multi-scale interaction but are less flexible than cross-attention mechanisms in dynamically adjusting modality-specific contribution weights for fine-grained feature alignment. Moreover, the retrieval efficacy of memory banks depends on the coverage of training data, which may struggle to capture rare pathology-genomics associations in historical records.